\ifdefined\XeTeXrevision\else\ifdefined\pdfoutput\pdfoutput=1\fi\fi
\PassOptionsToPackage{unicode}{hyperref}
\PassOptionsToPackage{hyphens}{url}
\PassOptionsToPackage{dvipsnames,svgnames,x11names}{xcolor}
\documentclass[
  10pt,
]{article}
\usepackage{xcolor}
\usepackage[margin=1in]{geometry}
\usepackage{amsmath,amssymb}
\usepackage{newunicodechar}
\newunicodechar{≈}{\ensuremath{\approx}}
\newunicodechar{≤}{\ensuremath{\leq}}
\newunicodechar{∈}{\ensuremath{\in}}
\usepackage{iftex}
\ifPDFTeX
  \usepackage[T1]{fontenc}
  \usepackage[utf8]{inputenc}
  \usepackage{textcomp} 
\else 
  \usepackage{unicode-math} 
  \defaultfontfeatures{Scale=MatchLowercase}
  \defaultfontfeatures[\rmfamily]{Ligatures=TeX,Scale=1}
\fi
\usepackage{lmodern}
\ifPDFTeX\else
\fi
\IfFileExists{upquote.sty}{\usepackage{upquote}}{}
\IfFileExists{microtype.sty}{
  \usepackage[]{microtype}
  \UseMicrotypeSet[protrusion]{basicmath} 
}{}
\makeatletter
\@ifundefined{KOMAClassName}{
  \IfFileExists{parskip.sty}{%
    \usepackage{parskip}
  }{
    \setlength{\parindent}{0pt}
    \setlength{\parskip}{6pt plus 2pt minus 1pt}}
}{
  \KOMAoptions{parskip=half}}
\makeatother
\usepackage{longtable,booktabs,array}
\usepackage{calc} 
\usepackage{etoolbox}
\makeatletter
\patchcmd\longtable{\par}{\if@noskipsec\mbox{}\fi\par}{}{}
\makeatother
\IfFileExists{footnotehyper.sty}{\usepackage{footnotehyper}}{\usepackage{footnote}}
\makesavenoteenv{longtable}
\usepackage{graphicx}
\makeatletter
\newsavebox\pandoc@box
\newcommand*\pandocbounded[1]{
  \sbox\pandoc@box{#1}%
  \Gscale@div\@tempa{\textheight}{\dimexpr\ht\pandoc@box+\dp\pandoc@box\relax}%
  \Gscale@div\@tempb{\linewidth}{\wd\pandoc@box}%
  \ifdim\@tempb\p@<\@tempa\p@\let\@tempa\@tempb\fi
  \ifdim\@tempa\p@<\p@\scalebox{\@tempa}{\usebox\pandoc@box}%
  \else\usebox{\pandoc@box}%
  \fi%
}
\def\fps@figure{htbp}
\makeatother
\providecommand{\tightlist}{%
  \setlength{\itemsep}{0pt}\setlength{\parskip}{0pt}}
\usepackage{bookmark}
\IfFileExists{xurl.sty}{\usepackage{xurl}}{} 
\hypersetup{
  colorlinks=true,
  linkcolor={Maroon},
  filecolor={Maroon},
  citecolor={Blue},
  urlcolor={Blue},
  pdfcreator={LaTeX via pandoc}}

\author{}
\date{}

\begin{document}

\title{Abliteration Is Not a Scalpel:\\ Off-Target Effects of Refusal Removal on Decision Disposition Across Model Families}
\author{Aleksander Fafuła, PhD\\ independent researcher \textbullet\ \href{https://funduszai.pl}{funduszai.pl} \textbullet\ \href{mailto:aleksander@fafula.com}{aleksander@fafula.com}}
\date{Draft v0.5 --- 2026-07-19}
\maketitle

\begin{abstract}

Abliteration --- deleting a model\textquotesingle s refusal direction
from its weights --- is the standard recipe behind popular "uncensored"
open-weight models. It is advertised as surgical: remove refusal, change
nothing else. We show the surgery is not clean. As a disposition probe
we use 21,600 decisions under uncertainty --- weekly up/down calls on 60
Warsaw Stock Exchange equities over 18 weeks, replayed through a frozen
multi-agent pipeline so the decision-layer model is the only variable (5
samples per cell, zero generation failures). The task elicits no
refusals at all --- base arms completed all 10,800 of their decisions
without refusing once --- so there was no refusal behavior to remove,
and any between-arm delta is pure side effect. Holding provenance
constant --- official BF16 checkpoints, a single abliteration author
(huihui-ai), an identical serving stack, one byte-identical frozen
prompt --- we compare base and abliterated arms of two
Mixture-of-Experts families (Gemma-4-26B-A4B-it and
Qwen3-30B-A3B-Instruct-2507). Three effects replicate across both
families (weeks-clustered bootstrap CIs excluding zero): abliterated
models are systematically \textbf{more optimistic} (+12.2 pp Gemma, +7.4
pp Qwen more up-bets; the confirmed preregistered endpoint),
\textbf{justify themselves at greater length}, and use \textbf{fewer
explicit uncertainty words} with more concessive rhetoric in their
self-critiques (both exploratory; length was preregistered as a
covariate, not a directional hypothesis). A fourth effect \textbf{reverses sign}: the same
operation makes Gemma-abliterated \emph{less} confident and
Qwen-abliterated \emph{more} (family CIs non-overlapping) --- one weight
surgery, opposite shifts in expressed confidence. A capability-covariate
analysis rules out instruction-following degradation as the driver, and
no arm shows economic skill --- the apparent edge of abliterated arms is
regime beta, not alpha. Along the way our provenance audit caught two
independent contamination channels (a mismatched-quantizer pilot pair,
whose headline finding the clean design overturns, and a stale chat
template in a community checkpoint that silently mangled the rendered
prompt) --- evidence that in studies of community-modified checkpoints,
toolchain artifacts are the rule, not the exception.
Abliteration\textquotesingle s off-target footprint is real, partially
universal, and model-dependent in direction: whoever deploys an
"uncensored" model as an agent is deploying a measurably different
decision-maker, not the base model minus refusals.

\end{abstract}

\paragraph{Key findings}

\begin{enumerate}
\def\labelenumi{\arabic{enumi}.}
\tightlist
\item
  \textbf{Removing refusals adds optimism.} On identical evidence,
  abliterated models bet on the upside +7.4 to +12.2 pp more often ---
  in both families and on both market segments (preregistered P1).
\item
  \textbf{Same surgery, opposite confidence shifts.} Abliteration makes
  Gemma \emph{less} confident and Qwen \emph{more} --- non-overlapping
  CIs. Off-target effects couple to model internals; "disinhibition"
  predicts the wrong pattern.
\item
  \textbf{The vocabulary of doubt thins everywhere.} Both abliterated
  arms use fewer explicit uncertainty words and more concessions
  ("although\ldots") in their forced self-critiques --- while stance
  stability under resampling is unchanged in every arm. Expressed doubt
  lives in the report, not in the behavior.
\item
  \textbf{No skill anywhere.} Every arm is a coin flip against the
  market; the abliterated arms\textquotesingle{} apparent trading edge
  flips sign with the market regime --- amplified beta, not
  intelligence.
\item
  \textbf{Two contamination channels caught.} A quantizer mismatch and a
  stale community chat template each produced publishable-looking
  artifacts; token-level provenance auditing killed both.
\end{enumerate}

\begin{center}\rule{0.5\linewidth}{0.5pt}\end{center}

\subsection{1. Introduction}\label{1-introduction}

Open-weight "uncensored" models are frequently produced by
\textbf{abliteration} (Arditi et al., 2024; Labonne, 2024): identifying
a refusal direction in activation space and orthogonalizing the weights
so the model no longer expresses it. The stated goal is surgical ---
remove refusals, change nothing else. Whether that surgery is clean is a
safety question: a model whose \emph{disposition} (its priors over risk,
optimism, decisiveness) shifts as a side effect could behave
unexpectedly in agentic deployments far from any refusal. The concern is
a behavioral cousin of emergent misalignment (Betley et al., 2025): a
narrow weight-space intervention producing broad, unintended shifts.

Measuring "disposition" is hard because most benchmarks probe
capability, not inclination. We use a task engineered to elicit a latent
stance under uncertainty: a fund\textquotesingle s \textbf{Research
Director} persona issues a weekly directional call (up or down) on a
stock, given a fixed bundle of analyst briefs and a debate transcript.
The task is deliberately chosen where prediction is near-hopeless ---
the production system it derives from frames weekly direction as "close
to a coin flip" --- so any systematic difference between arms is a
difference in \emph{inclination}, not in an (absent) predictive edge.
Because the upstream is frozen and identical across model arms, the only
variable is the decision-layer model. \textbf{The financial framing is
instrumental, not the point}: we care about the distribution of stances,
confidence, horizon, and self-justification the model produces, as a
readout of disposition. Throughout, we report the share of up-bets as
the \textbf{optimism rate} (the trading term is \emph{bull rate}); "more
optimistic" means the model bets on the upside more often given
identical evidence.

Our contributions:

\begin{enumerate}
\def\labelenumi{\arabic{enumi}.}
\tightlist
\item
  A \textbf{clean-provenance, cross-family} design isolating
  abliteration as the sole variable, with a preregistered frozen prompt
  and 21,600 decisions.
\item
  Evidence that abliteration\textquotesingle s off-target effects are
  \textbf{partially universal}: a directional optimism shift
  (preregistered) generalizes across two families, as do ---
  exploratorily --- self-justification verbosity and a lexical thinning
  of uncertainty vocabulary; expressed confidence does \textbf{not}
  generalize --- it reverses sign.
\item
  A \textbf{capability-covariate} control showing the disposition shifts
  are not explained by instruction-following degradation.
\item
  A \textbf{provenance-correction} result: a prior "excision of doubt"
  finding is shown to be an artifact of a contaminated
  (mismatched-quantizer) pilot pair.
\item
  A fully disclosed \textbf{preregistration accounting}, including
  endpoints that came out degenerate or opposite to the preregistered
  hypothesis (§3.5).
\end{enumerate}

\subsection{2. Related Work}\label{2-related-work}

\textbf{Refusal directions and abliteration.} Arditi et al. (2024) show
refusal in open-weight chat models is mediated by a single activation
direction and that weight orthogonalization removes refusal with
(reportedly) minimal capability cost; Labonne (2024) popularized the
recipe as "abliteration," and huihui-ai maintains the largest public
catalog of abliterated checkpoints --- the supply side of the ecosystem
we test. Our question is precisely the footprint that capability
benchmarks in these works do not measure: disposition.

\textbf{Steering and side effects of weight interventions.} Activation
addition and contrastive steering (Turner et al., 2023; Panickssery et
al., 2024) established that low-rank directions carry behavioral traits.
Fine-tuning is known to degrade alignment even unintentionally (Qi et
al., 2023), and narrow fine-tuning can produce broad misalignment
(Betley et al., 2025). We extend this line to \emph{inference-free
weight surgery} and to a behavioral readout far from the
intervention\textquotesingle s target.

\textbf{Verbalized confidence and LLM behavioral economics.}
LLMs\textquotesingle{} stated confidence is measurable and manipulable
(Kadavath et al., 2022; Tian et al., 2023); LLMs are increasingly used
as simulated economic agents with systematic biases (Horton, 2023). We
combine the two: expressed confidence and stance as psychometrics of a
weight-level intervention.

\textbf{Toolchain confounds.} Compression and quantization measurably
change model behavior beyond benchmark scores (Jaiswal et al., 2023).
Our pilot-correction result (§5.3) is a concrete instance: a disposition
finding that flipped once quantizer provenance was cleaned.

\subsection{3. Method}\label{3-method}

\subsubsection{3.1 Task and replay
harness}\label{31-task-and-replay-harness}

We reuse the decision layer of a production multi-agent equity pipeline
(\emph{funduszai.pl} --- the author\textquotesingle s non-commercial,
fully transparent AI fund simulator on the Warsaw Stock Exchange, which
frames weekly direction as "close to a coin flip" and studies model
reasoning rather than prediction). The pipeline runs three tiers: four
neutral \textbf{analysts} (news, technicals, fundamentals, macro), eight
\textbf{experts} in forced bull/bear debate pairs, and a board stage.
The upstream (four analyst briefs + debate transcript + context bundle)
is read \textbf{read-only} from the production store and frozen per
cell; only the \textbf{Research-Director} decision is regenerated by the
model under test. We deliberately drop the downstream CEO stage
(manager-only) to minimize stochastic compounding and isolate a single
decision.

The manager emits a single JSON object:
\texttt{stance\ ∈\ \{bull,\ bear\}} (an up/down bet),
\texttt{time\_horizon\ ∈\ \{1w,2w,4w\}}, a \texttt{thesis}, 2--4
\texttt{key\_risks}, a \texttt{confession} (one sentence of truth
\emph{against} its own decision), a \texttt{reversal\_trigger} (the
concrete event that would flip the call), and two self-assessments
\texttt{confidence} and \texttt{p\_wrong\ ∈\ {[}0,1{]}}. The confession
and reversal\_trigger fields are deliberate probes of expressed doubt.
The prompt is frozen (sha-256 recorded) and byte-identical across all
four model arms.

\subsubsection{3.2 Models and provenance
control}\label{32-models-and-provenance-control}

Two MoE families, each as a base/abliterated pair: the base is the
official publisher checkpoint (Google, Alibaba), the abliterated
counterpart comes from a \textbf{single abliteration author (huihui-ai)}
for both families. All four arms are served in \textbf{BF16 exactly as
published} --- no re-quantization, no intermediate conversions ---
through an identical vLLM stack (vLLM 0.24.0, same flags, port, context
length 16,384), differing in weights and a distinguishable alias:

{\def\LTcaptype{none} 
\small
\begin{longtable}[]{@{}lll@{}}
\toprule\noalign{}
Family & arm & HF repo \\
\midrule\noalign{}
\endhead
\bottomrule\noalign{}
\endlastfoot
Gemma-4-26B-A4B-it & base & \texttt{google/gemma-4-26B-A4B-it} \\
& abliterated & \texttt{huihui-ai/Huihui-gemma-4-26B-A4B-it-abliterated} \\
Qwen3-30B-A3B-Instruct-2507 & base & \texttt{Qwen/Qwen3-30B-A3B-Instruct-2507} \\
& abliterated & \texttt{huihui-ai/Huihui-Qwen3-30B-A3B-Instruct-2507-abliterated} \\
\end{longtable}
}

Holding the abliteration author fixed isolates the \emph{method} and
places the cross-family contrast on the \emph{base model}. Instruct-2507
is non-thinking, avoiding reasoning leakage into JSON. Arm labels are
sanity-checked with a small label-integrity battery (6 sensitive
prompts, PL/EN): base arms refuse at or near ceiling, abliterated arms
at 0--17\%. We treat this as a label check, not an estimate of refusal
rates. A prior pilot used an NVFP4 pair with \textbf{mismatched
quantizers} (ModelOpt W4A4 vs llmcompressor, plus a post-hoc router
splice and a locally patched chat template); we treat that pilot as
contaminated and rebuild from clean BF16 checkpoints (see §5.3).

\textbf{Provenance audit (full report in \texttt{PROVENANCE.md}).} We
hashed every weight shard, config, tokenizer, and chat template of all
four checkpoints, diffed each pair, and verified prompt rendering
token-for-token against the logged per-request token counts. Qwen is
fully clean: configs identical, tokenizer byte-identical, and although
the abliterated repo ships a syntactically different chat template, it
renders byte-identically for our message shape --- input token counts
match in 1,080/1,080 cells across arms. For Gemma the audit caught a
real serving artifact: the huihui repo ships an \textbf{older
\texttt{chat\_template.jinja}} that mishandles structured message
content, so in an initial run the abliterated arm\textquotesingle s
system prompt was rendered wrapped in a list literal with escaped
newlines
(\texttt{{[}\{\textquotesingle{}type\textquotesingle{}:\ \textquotesingle{}text\textquotesingle{},\ \textquotesingle{}text\textquotesingle{}:\ \textquotesingle{}…\textbackslash{}\textbackslash{}n…\textquotesingle{}\}{]}}).
Rather than argue the mangling harmless, we \textbf{discarded that run
entirely} and re-collected the full Gemma abliterated arm with the base
repo\textquotesingle s chat template forced at serving time (weights
unchanged; rendering then verified byte-identical to the base arm,
token-for-token on logged counts). All Gemma numbers reported in this
paper come exclusively from the clean rerun. The episode itself is a
finding: community abliterated checkpoints can differ from their base
not only in weights but in serving-relevant metadata, and studies that
do not verify rendered prompts token-for-token can silently measure the
template, not the intervention (cf. §5.3).

\subsubsection{3.3 Grid and metrics}\label{33-grid-and-metrics}

Universe: WIG20 (20 large caps) + mWIG40 (40 mid caps) = 60 stocks.
Window: 18 weekly Saturdays, 2026-03-07\ldots2026-07-04. Market context,
stated up front because every level (though no delta) depends on it: the
WIG20 benchmark gained +15.4\% over the 18 economically settled weeks,
12 benchmark-positive weeks to 6 negative --- an up-market-heavy but
two-sided window (both regimes are used in the split of §4.6). Samples:
n=5 per cell (temperature 1.0, top-p 0.95, top-k 64, identical across
arms). Total: 5,400 decisions per arm × 4 arms = \textbf{21,600}, with
zero generation failures. The harness allowed up to 3 content retries
per decision (malformed JSON, out-of-range fields); this mechanism was
\textbf{never triggered} in any arm --- the only retries in the run logs
of the four analyzed arms are three transient network timeouts (all in
base-arm runs). In particular, no arm ever refused the task: the "no
refusal behavior to remove" claim (Abstract, §5.1) holds at the level of
first attempts, not just final outputs.

Disposition metrics per arm: optimism rate (share of up-bets);
confidence distribution (mean and low-tail mass); horizon collapse
(fraction 1w); confession length and lexical doubt-score; cross-sample
unanimity (flip-rate); \texttt{p\_wrong} coherence. Economic metrics:
weekly PnL (Monday open → Friday close of the week after the decision),
hit-rate, signed return, and a regime split on the benchmark sign.

\subsubsection{3.4 Statistics and
preregistration}\label{34-statistics-and-preregistration}

The unit of independence is the \textbf{week} (multiple samples per cell
buy precision, not independent bets). We report cluster bootstrap 95\%
CIs resampling the 18 weeks (N=5,000). Three primary hypotheses were
preregistered with the frozen prompt before any grid data
(\texttt{PREREGISTRATION-v2.md}, tag \texttt{freeze-v2-2026-07-05}): P1
optimism shift (abl \textgreater{} base), P2 low-confidence tail (base
\textgreater{} abl), P3 confession doubt-score (base \textgreater{}
abl). All other analyses are exploratory and labeled as such; with three
preregistered primaries the multiple-comparisons surface is small, and
the exploratory findings should be read as hypothesis-generating. The
design was \textbf{phased by preregistration}: WIG20 was Phase 1; the
mWIG40 run was gated on Phase 1 showing any non-null primary endpoint,
and then serves as a within-family replication on a fresh segment. The
gate fired (P1 strongly non-null), so mWIG40 was run in full. We report
capability covariates per arm to guard against a
compliance-vs-disposition confound.

\subsubsection{3.5 Deviations from
preregistration}\label{35-deviations-from-preregistration}

We disclose every deviation:

\begin{itemize}
\tightlist
\item
  \textbf{P2 threshold.} The preregistration defined the low-confidence
  tail as \texttt{confidence\ ≤\ 0.45}. That tail turned out to be
  \textbf{degenerate in all four arms} (≤0.2\% of decisions everywhere):
  in Gemma it is identically zero in both arms (Δ = 0.00 pp); in Qwen Δ
  = −0.15 pp, CI {[}−0.24, −0.06{]} (the preregistered sign, but
  negligible in magnitude). \textbf{P2 as preregistered is therefore not
  confirmed.} The ≤0.60 tail used in §4 is an exploratory
  re-parameterization chosen after seeing the data; at that threshold
  the Gemma effect is \emph{opposite} to the preregistered direction
  (the abliterated arm\textquotesingle s low tail grows). We report
  both.
\item
  \textbf{mWIG40 gate timing.} The preregistered gate (run mWIG40 only
  if Phase 1 shows a non-null primary endpoint) was originally evaluated
  on Phase-1 data that included the Gemma abliterated run later
  discarded for the chat-template artifact (§3.2). The gate criterion
  also holds, more strongly, in the clean re-collection (P1 = +14.3 pp
  on WIG20), so the gating decision is unaffected.
\item
  \textbf{P3 lexicon timing.} The preregistration required the doubt
  lexicon to be frozen in \texttt{analysis/}\allowbreak\texttt{doubt\_lexicon\_v2.txt}
  \emph{before} the run. The file was not created before the run; the
  lexicon was frozen post-hoc (2026-07-06). P3 is therefore reported as
  \textbf{exploratory}, with a sensitivity table across lexicon variants
  (§4.4).
\item
  \textbf{Aggregation unit.} The preregistration described aggregating
  n=5 samples to a per-cell consensus before testing, with week×stock as
  the unit. The reported analyses use decision-level data with the
  bootstrap clustered at the coarser \textbf{week} level, which is the
  more conservative choice (it allows arbitrary within-week correlation
  across stocks and samples).
\item
  \textbf{Window start date.} The preregistration wrote 2026-03-01; the
  first Saturday in the data is 2026-03-07 (2026-03-01 was a Sunday ---
  a typo). The window is the 18 Saturdays as specified.
\end{itemize}

\subsection{4. Results}\label{4-results}

\subsubsection{4.1 Disposition deltas (abl − base) with weeks-clustered
bootstrap
CIs}\label{41-disposition-deltas-abl--base-with-weeks-clustered-bootstrap-cis}

{\def\LTcaptype{none} 
\begin{longtable}[]{@{}lllll@{}}
\toprule\noalign{}
Family & metric & Δ (abl−base) & 95\% CI & interpretation \\
\midrule\noalign{}
\endhead
\bottomrule\noalign{}
\endlastfoot
Gemma & optimism rate & \textbf{+12.2 pp} & {[}+8.5, +16.0{]} & more
optimistic \\
Gemma & mean confidence & \textbf{−0.008} & {[}−0.010, −0.006{]} &
\textbf{less} confident \\
Gemma & confession length & +4.0 words & {[}+3.8, +4.2{]} & more
verbose \\
Gemma & horizon (1w share) & −0.24 pp & {[}−0.54, −0.02{]} & both ≈100\%
(nil) \\
Qwen & optimism rate & \textbf{+7.4 pp} & {[}+5.0, +9.7{]} & more
optimistic \\
Qwen & mean confidence & \textbf{+0.109} & {[}+0.097, +0.120{]} &
\textbf{more} confident \\
Qwen & confession length & +7.4 words & {[}+7.1, +7.7{]} & more
verbose \\
Qwen & horizon (1w share) & \textbf{−42.0 pp} & {[}−46.5, −37.5{]} &
opens the horizon \\
\end{longtable}
}

All deltas exclude zero (\textbf{Figure 1}). The deltas are
\textbf{segment-stable}: WIG20 and mWIG40 yield near-identical values
within each family (e.g. Gemma optimism shift +14.3 pp WIG20 / +11.1 pp
mWIG40; Qwen +5.7 / +8.2), so no effect is a large-cap artifact.

\pandocbounded{\includegraphics[width=\linewidth,keepaspectratio]{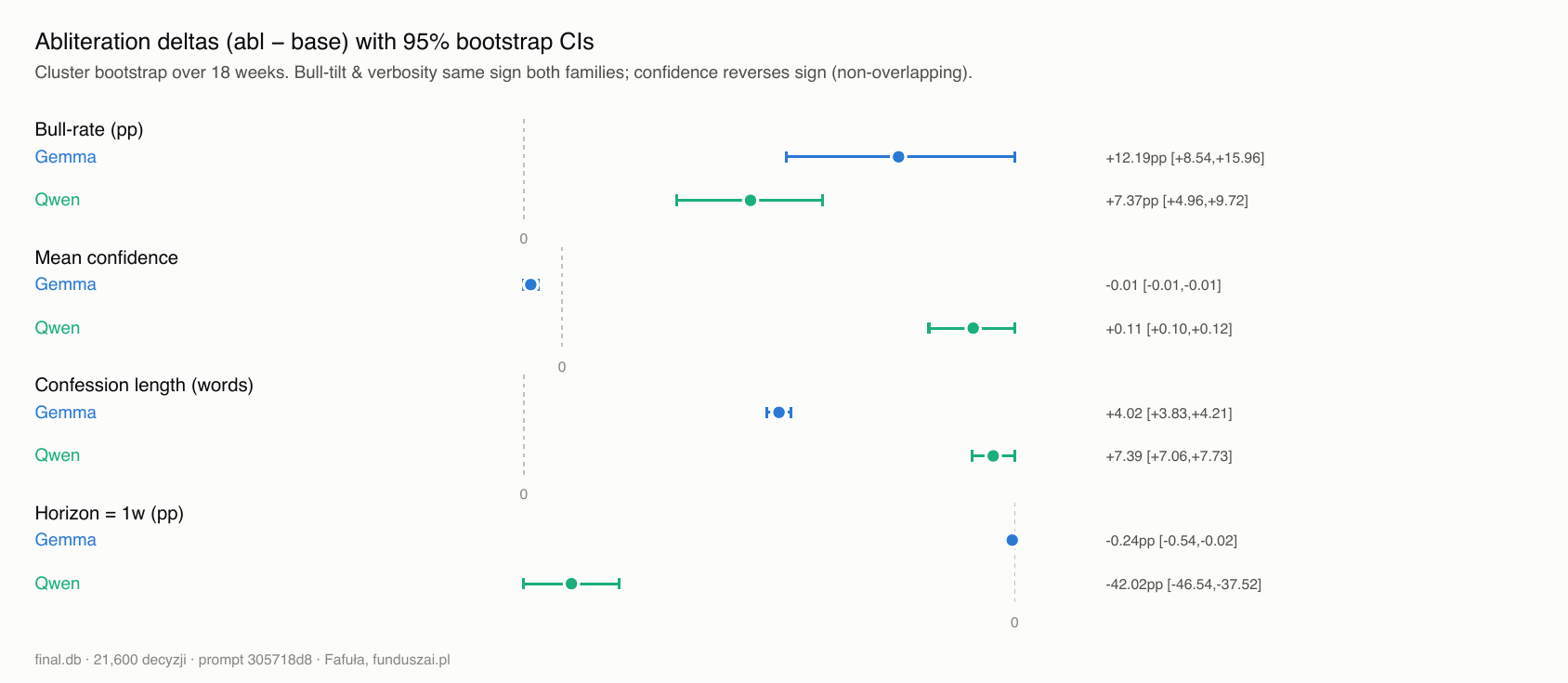}}
\emph{Figure 1. Abliteration deltas (abl − base) with 95\% bootstrap CIs
(weeks as clusters). Optimism shift and confession verbosity share sign
across families; mean confidence reverses sign, and the family CIs do
not overlap.}

\subsubsection{4.2 What generalizes and what does
not}\label{42-what-generalizes-and-what-does-not}

\begin{itemize}
\tightlist
\item
  \textbf{Generalizes (same sign, both families):} the \textbf{optimism
  shift} (abliterated models bet on the upside more often everywhere;
  stronger in Gemma, whose base is more neutral, than in Qwen, whose
  base is already a strong regime-follower) and
  \textbf{self-justification verbosity} (abliterated models explain
  themselves at greater length).
\item
  \textbf{Reverses sign (model-dependent):} \textbf{expressed
  confidence}. Gemma-abliterated is \emph{less} confident (mean −0.008;
  its ≤0.60 tail grows from 16\% to 30\%); Qwen-abliterated is
  \emph{more} confident (mean +0.109). The Gemma effect is small in
  magnitude but its CI excludes zero, and the two family CIs are
  \textbf{non-overlapping} (Gemma entirely negative, Qwen entirely
  positive), so this is not sampling noise. Caveat: Qwen operates in a
  saturated high-confidence regime (≈0\% low tail in both arms), so the
  tail metric is degenerate for Qwen and the reversal is read from the
  mean. Because both arms of a pair share the base
  model\textquotesingle s verbal number habits, the within-pair delta
  cannot be explained by family-specific confidence-token conventions
  --- those cancel in the contrast.
\item
  \textbf{Family-specific:} the \textbf{horizon} effect.
  Qwen-abliterated \emph{opens} the horizon distribution (base ≈95\%
  collapse to 1w → abliterated ≈55\%), whereas both Gemma arms are
  ≈fully collapsed at 1w (99.6--99.9\%, no room to move).
\end{itemize}

\pandocbounded{\includegraphics[width=\linewidth,keepaspectratio]{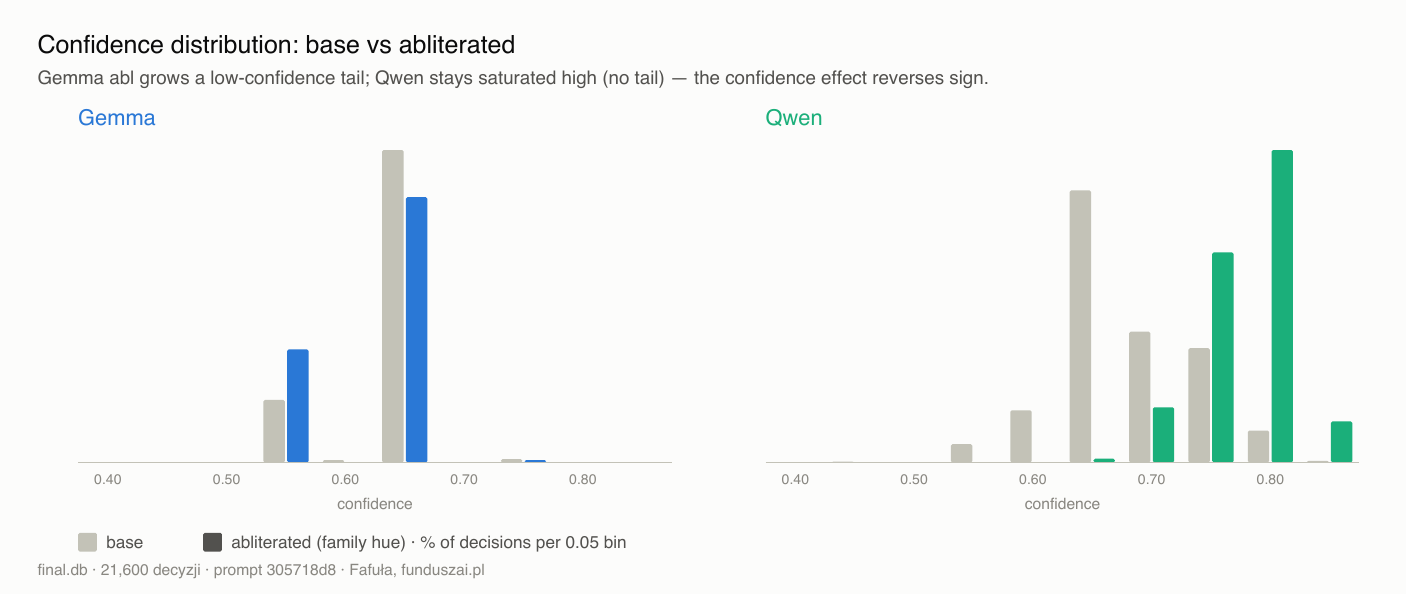}}
\emph{Figure 2. Confidence distributions per arm. Gemma-abliterated
shifts mass toward lower confidence; Qwen-abliterated shifts higher ---
the sign reversal.}

\pandocbounded{\includegraphics[width=\linewidth,keepaspectratio]{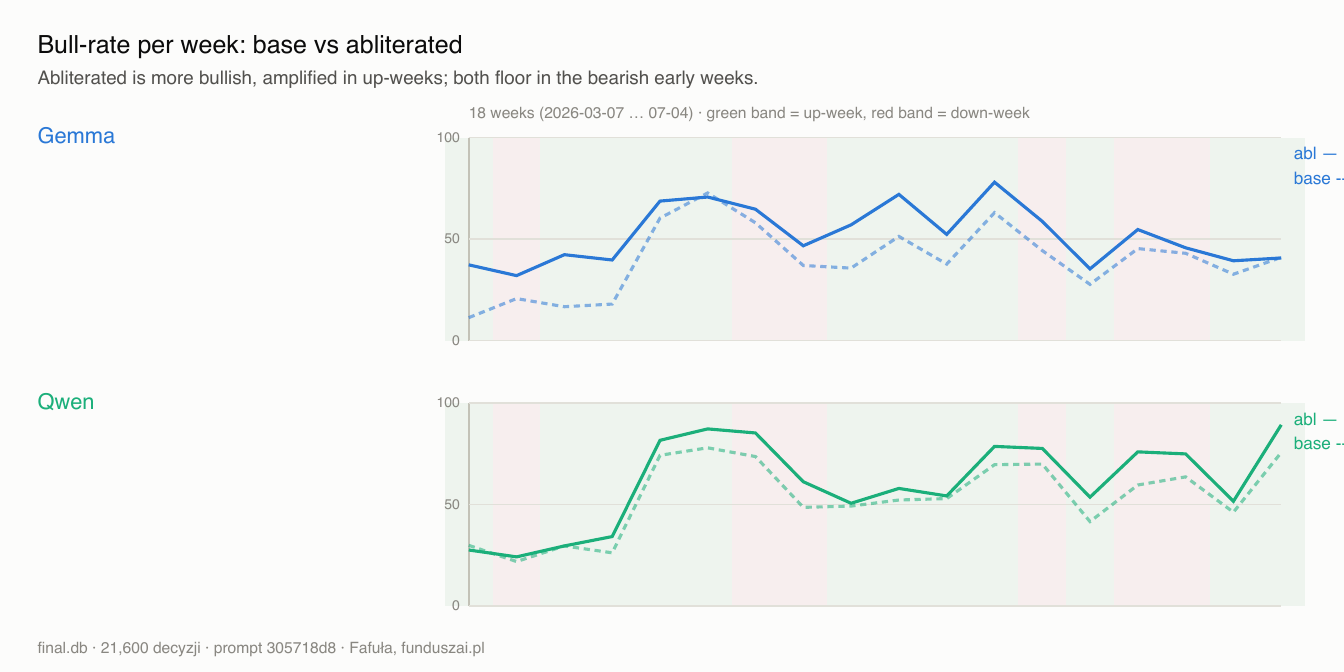}}
\emph{Figure 3. Optimism rate per week and arm. The abliterated arm sits
above its base in both families, with regime amplification in up-weeks.}

\pandocbounded{\includegraphics[width=\linewidth,keepaspectratio]{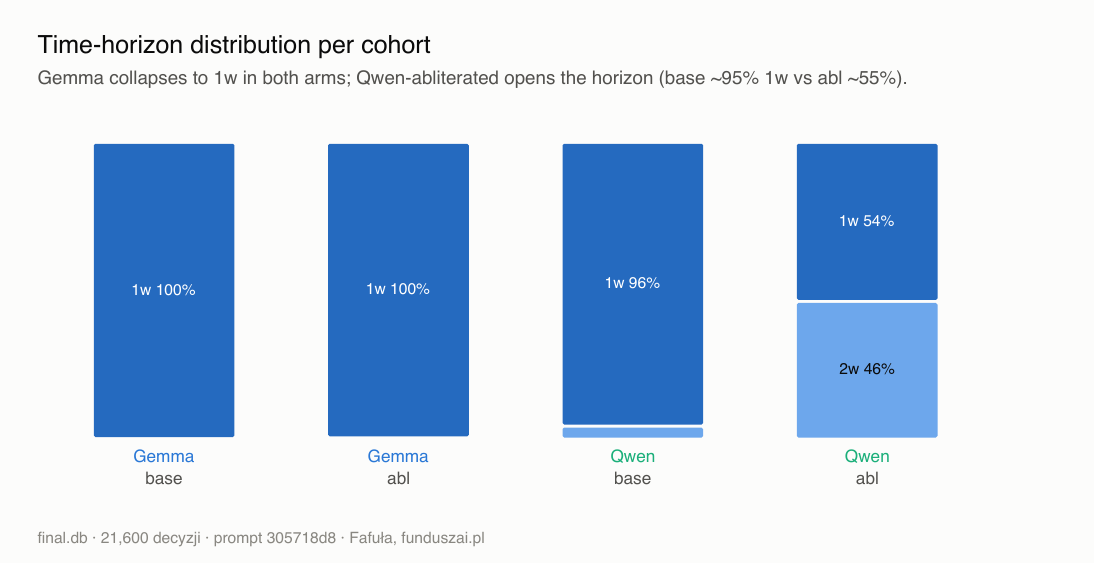}}
\emph{Figure 4. Horizon distributions: Qwen-abliterated opens the
horizon (≈95\% → ≈55\% at 1w); both Gemma arms are collapsed at 1w.}

\subsubsection{4.3 Capability covariates rule out a compliance
confound}\label{43-capability-covariates-rule-out-a-compliance-confound}

{\def\LTcaptype{none} 
\begin{longtable}[]{@{}lllllll@{}}
\toprule\noalign{}
Family & arm & JSON valid & \#risks & risks∈{[}2,4{]} & reversal
specific & confession generic \\
\midrule\noalign{}
\endhead
\bottomrule\noalign{}
\endlastfoot
Gemma & base & 100\% & 3.11 & 100\% & 95\% & 0\% \\
Gemma & abl & 100\% & 2.98 & 97\% & 85\% & 0\% \\
Qwen & base & 100\% & 4.00 & 100\% & 100\% & 0.1\% \\
Qwen & abl & 100\% & 4.00 & 100\% & 100\% & 0.1\% \\
\end{longtable}
}

Instruction-following is preserved. For \textbf{Qwen it is identical
across arms}, yet disposition shifts are large --- so the disposition
deltas are not a byproduct of capability loss. The one exception is a
modest drop in Gemma-abliterated \texttt{reversal\_trigger} specificity
(95\%→85\% contain a concrete price/level), noted as a limitation but
insufficient to explain the primary effects.

\subsubsection{4.4 Expressed doubt: the number and the vocabulary move
separately}\label{44-expressed-doubt-the-number-and-the-vocabulary-move-separately}

The preregistered P2 (low-confidence tail at ≤0.45) came out degenerate
in all arms (§3.5). The exploratory picture is richer. On the confidence
\emph{number}, the families diverge (§4.2). On the confession
\emph{vocabulary} (doubt-score: lexicon tokens per 100 words; lexicon
frozen post-hoc, so exploratory), the families \textbf{agree}:

{\def\LTcaptype{none} 
\begin{longtable}[]{@{}llllll@{}}
\toprule\noalign{}
Family & lexicon variant & base & abl & Δ (abl−base) & 95\% CI \\
\midrule\noalign{}
\endhead
\bottomrule\noalign{}
\endlastfoot
Gemma & uncertainty (CORE) & 3.68 & 2.73 & \textbf{−0.95} & {[}−1.11,
−0.80{]} \\
Gemma & concessive & 0.79 & 2.18 & \textbf{+1.39} & {[}+1.31,
+1.48{]} \\
Gemma & combined (FULL) & 4.47 & 4.90 & +0.43 & {[}+0.27, +0.58{]} \\
Qwen & uncertainty (CORE) & 3.46 & 1.29 & \textbf{−2.17} & {[}−2.25,
−2.09{]} \\
Qwen & concessive & 1.29 & 2.28 & \textbf{+1.00} & {[}+0.88, +1.12{]} \\
Qwen & combined (FULL) & 4.74 & 3.57 & −1.17 & {[}−1.32, −1.03{]} \\
\end{longtable}
}

In \textbf{both families}, abliterated arms use \textbf{fewer explicit
uncertainty markers} (risk/doubt/uncertainty words; a minimal 3-prefix
variant agrees) and \textbf{more concessive rhetoric}
("although/despite/but"), per 100 words of confession. The combined
score depends on the mix --- which is exactly why the lexicon-variant
sensitivity is reported. Read together with §4.2: abliteration thins the
\emph{vocabulary of doubt} in both families (the direction the pilot
called "excision of doubt"), while the \emph{stated confidence number}
moves in family-dependent directions. Expressed doubt is not one channel
but at least two, and abliteration decouples them.

\subsubsection{4.5 Stated versus enacted
uncertainty}\label{45-stated-versus-enacted-uncertainty}

Cross-sample unanimity is 91--95\% in all four clean arms (flip-rate
5--9\%): stance is highly stable under resampling. The contaminated
pilot showed 68\% flip at N=10 (45\% at N=5, apples-to-apples) --- its
instability was largely an artifact of prompt complexity and W4A4
activation noise, not genuine epistemic uncertainty. The dissociation
matters beyond methodology: Gemma-abliterated \emph{declares} more
uncertainty (lower confidence numbers) yet \emph{acts} just as
decisively (no extra stance flipping). Expressed uncertainty lives in
the report, not in the behavior --- a caution for anyone using stated
confidence as a proxy for an agent\textquotesingle s actual
decisiveness.

\subsubsection{4.6 No economic skill; the edge is
beta}\label{46-no-economic-skill-the-edge-is-beta}

No arm beats a buy-and-hold benchmark. For Gemma, the abliterated
arm\textquotesingle s apparent PnL edge is \textbf{regime beta}: in a
weeks-clustered regime split, its advantage flips sign (WIG20 up-weeks
+0.57\% vs base −0.04\%; down-weeks −0.61\% vs +0.05\%), while the base
arm is regime-neutral. For Qwen, the abliterated−base PnL difference is
negligible, because the Qwen base is \emph{already} a strong
regime-follower, leaving little marginal beta to add (\textbf{Figure
5}). Overall: abliteration changes \textbf{disposition, not skill}.

\pandocbounded{\includegraphics[width=\linewidth,keepaspectratio]{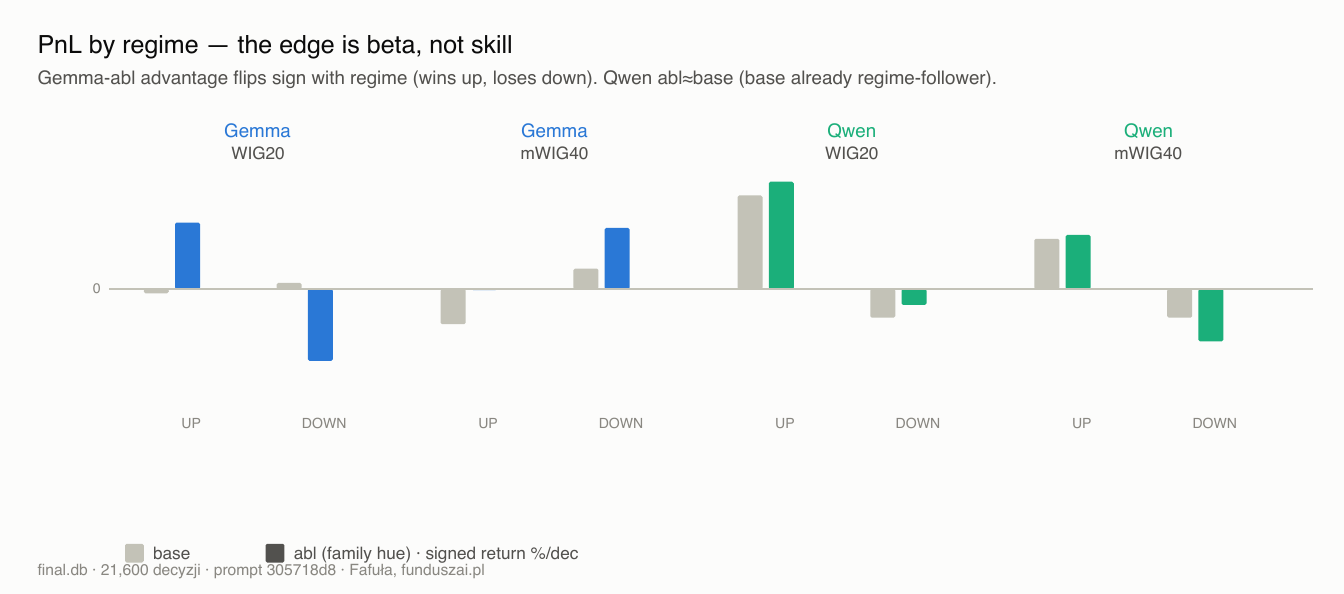}}
\emph{Figure 5. Signed weekly return by market regime (up/down weeks).
The abliterated arms\textquotesingle{} apparent edge flips sign with the
regime --- beta, not skill.} Economic claims are low-powered by
construction (≈18 independent weeks) and we make none beyond this null.

\subsection{5. Discussion}\label{5-discussion}

\subsubsection{5.1 Abliteration\textquotesingle s off-target footprint
is real but only partly
universal}\label{51-abliterations-off-target-footprint-is-real-but-only-partly-universal}

One preregistered effect (the directional optimism shift, P1) and two
exploratory effects (self-justification verbosity --- preregistered only
as a reported covariate, not a directional hypothesis --- and the
thinning of uncertainty vocabulary) generalize across two families
abliterated by the same method; one effect (expressed confidence)
reverses sign.
Because the abliteration \emph{method} is held fixed (huihui-ai), the
reversal localizes the variability to the \textbf{base model} --- the
same refusal-direction removal interacts with each
model\textquotesingle s representations differently. This is the
paper\textquotesingle s central, and most surprising, claim.

Two features of the design rule out the deflationary reading that
"uncensored models simply speak more freely." First, the probe task
contains nothing to censor: base arms never refused a single one of
their 10,800 decisions, so no refusal behavior was suppressing an
underlying stance that abliteration could have "released." Second, a
disinhibition account predicts a uniform direction --- more confidence
everywhere --- whereas the observed confidence shift reverses sign
across families. What remains is not liberation but collateral damage:
an intervention aimed at one behavior perturbing an unrelated decision
mechanism, with model-dependent direction.

\subsubsection{5.2 A mechanistic hook}\label{52-a-mechanistic-hook}

Why would removing the same refusal direction make one model less
confident and another more? A natural follow-up projects the refusal
direction onto a disposition axis (optimism/caution) in each
model\textquotesingle s activation space and asks whether the induced
shift\textquotesingle s sign matches the behavioral reversal. We leave
this to future work but note our behavioral result predicts a
model-dependent geometric relationship.

\subsubsection{5.3 Provenance correction of the
pilot}\label{53-provenance-correction-of-the-pilot}

A pilot on a mismatched-quantizer NVFP4 pair reported an "excision of
doubt" (abliterated less doubtful, low-confidence tail removed). On
clean BF16 checkpoints the picture inverts and splits: Gemma-abliterated
\emph{declares more} numeric doubt, and only the lexical channel (§4.4)
still moves in the pilot\textquotesingle s direction. More broadly, this
is a cautionary result: abliteration studies that do not control
quantizer/toolchain provenance may measure the quantizer, not the
intervention (cf. Jaiswal et al., 2023). Our own audit then caught a
second, independent instance within this study --- the abliterated Gemma
repo\textquotesingle s stale chat template mangling the rendered system
prompt (§3.2) --- detectable only because we verified rendered prompts
token-for-token against logged counts, and resolved by discarding and
re-collecting the affected arm. Two contamination channels in one
research program suggest such artifacts are the rule, not the exception,
in community-checkpoint studies.

\subsubsection{5.4 Anticipated
objections}\label{54-anticipated-objections}

\textbf{"Compared to what?"} Every number in this paper is a within-pair
contrast: a base model versus its own abliterated derivative, on
byte-identical inputs --- the same 60 stocks, the same 18 weeks, the
same frozen analyst briefs and debate transcripts, the same prompt, the
same serving stack, five samples per cell. We never compare Gemma to
Qwen directly, and no claim compares a model to the market. Everything
the two arms share --- news flow, market regime, prompt, pipeline ---
cancels in the difference.

\textbf{"The market was rising, so of course the model is bullish."}
Both arms saw exactly the same market. A rising market can lift both
arms\textquotesingle{} optimism \emph{levels}; it cannot open a
systematic \emph{gap} between two models reading identical evidence. The
gap is not a bull-market artifact: the optimism delta is positive in
down-weeks as well as up-weeks in both families (Gemma +13.8 pp up /
+9.0 pp down; Qwen +5.9 pp up / +10.3 pp down --- larger in down-weeks),
and it is positive in 16 of 18 individual weeks per family, with the
remaining weeks within −2.3 pp of zero.

\textbf{"With a different trend the result would change."} The
regime-dependent quantity is the \emph{level} of optimism --- and we
document that dependence ourselves (Qwen\textquotesingle s base is a
strong regime-follower, §4.2). The paper\textquotesingle s claim is the
\emph{delta}: abliterated arms sit above their own base on identical
evidence, in both regimes, on both segments. Where regime dependence
does bite --- trading performance --- the paper already says so: the
abliterated arms\textquotesingle{} apparent PnL edge is regime beta that
would flip sign in a bear market (§4.6). That is our finding, not our
blind spot.

\textbf{"18 weeks is too short."} For economic inference, yes --- which
is why the only economic claim in this paper is a null. The disposition
deltas rest on 21,600 decisions, with uncertainty quantified by the
deliberately conservative choice of clustering the bootstrap at the week
level (§3.4). A longer window would tighten the confidence intervals,
not conjure the effects: they are visible week by week.

\textbf{"Weekly stock direction is a coin flip, so the task is
meaningless."} The near-hopelessness is the design, not a defect.
Precisely because no model can win on skill, any stable between-arm
difference must be \emph{inclination}, not intelligence. Finance here is
an instrument: a generator of repeated, consequential, non-leakable
decisions under uncertainty with dated ground truth. The disposition
shifts it detects are properties of the model, and there is no reason to
expect them to stay confined to finance (§5.5).

\textbf{"Maybe abliteration just made the models worse."} Degradation
predicts noise and broken instruction-following. We observe the
opposite: capability covariates are preserved (Qwen\textquotesingle s
are \emph{identical} across arms while its disposition shifts are the
largest, §4.3), JSON validity is 100\% everywhere, and the shifts are
consistently signed across families and segments --- structure, not
damage.

\subsubsection{5.5 Limitations}\label{55-limitations}

Single abliteration author (huihui-ai) --- cross-family generalization
is across base models but one method; a second author (e.g. mlabonne)
would strengthen it, and huihui-ai\textquotesingle s exact procedure is
not fully public, which is itself an argument for the mechanistic
follow-up. Single task, single language (Polish), single
role-conditioned persona: deltas are shifts in a \emph{role}, not
necessarily a global trait. One decoding configuration (temperature 1.0,
top-p 0.95, top-k 64); a pilot temperature sweep suggested robustness,
but on the contaminated pair. A single 18-week, up-market-heavy window.
Two probes (\texttt{p\_wrong}, horizon in Gemma) are degenerate. The
optimism shift could reflect a more general positivity/agreeableness
shift rather than anything financial --- distinguishing the two needs a
non-financial disposition battery. No mechanistic grounding yet.

\subsection{6. Conclusion}\label{6-conclusion}

Abliteration is not a scalpel. Across two model families it reliably
shifts decision disposition toward optimism and longer
self-justification, thins the vocabulary of doubt, and --- depending on
the base model --- moves expressed confidence in opposite directions,
all under a fixed abliteration method, clean BF16 provenance, and a
frozen prompt. It confers no economic skill. The practical upshot is
simple: an "uncensored" model is not the base model minus refusals. It
is a different decision-maker --- one that bets the upside more, hedges
less in words while acting just as decisively, and whose confidence has
been silently re-tuned in a direction that cannot be predicted without
measuring it. These off-target, partially-universal effects are
safety-relevant for anyone deploying abliterated models as agents, and
they motivate a mechanistic account of how refusal-direction removal
couples to a model\textquotesingle s latent disposition.

\subsection{References}\label{references}

\begin{itemize}
\tightlist
\item
  Arditi, A., Obeso, O., Syed, A., Paleka, D., Panickssery, N., Gurnee,
  W., Nanda, N. (2024). \emph{Refusal in Language Models Is Mediated by
  a Single Direction.} arXiv:2406.11717.
\item
  Betley, J., Tan, D., Warncke, N., Sztyber-Betley, A., Bao, X., Soto,
  M., Labenz, N., Evans, O. (2025). \emph{Emergent Misalignment: Narrow
  Finetuning Can Produce Broadly Misaligned LLMs.} arXiv:2502.17424.
\item
  Horton, J. J. (2023). \emph{Large Language Models as Simulated
  Economic Agents: What Can We Learn from Homo Silicus?}
  arXiv:2301.07543.
\item
  Jaiswal, A., Gan, Z., Du, X., Zhang, B., Wang, Z., Yang, Y. (2023).
  \emph{Compressing LLMs: The Truth is Rarely Pure and Never Simple.}
  arXiv:2310.01382.
\item
  Kadavath, S., et al. (2022). \emph{Language Models (Mostly) Know What
  They Know.} arXiv:2207.05221.
\item
  Labonne, M. (2024). \emph{Uncensor any LLM with abliteration.} Hugging
  Face blog. \url{https://huggingface.co/blog/mlabonne/abliteration}
\item
  Panickssery, N., Gabrieli, N., Schulz, J., Tong, M., Hubinger, E.,
  Turner, A. M. (2024). \emph{Steering Llama 2 via Contrastive
  Activation Addition.} arXiv:2312.06681.
\item
  Qi, X., Zeng, Y., Xie, T., Chen, P.-Y., Jia, R., Mittal, P.,
  Henderson, P. (2023). \emph{Fine-tuning Aligned Language Models
  Compromises Safety, Even When Users Do Not Intend To!}
  arXiv:2310.03693.
\item
  Tian, K., Mitchell, E., Zhou, A., Sharma, A., Rafailov, R., Yao, H.,
  Finn, C., Manning, C. D. (2023). \emph{Just Ask for Calibration:
  Strategies for Eliciting Calibrated Confidence Scores from Language
  Models Fine-Tuned with Human Feedback.} arXiv:2305.14975.
\item
  Turner, A. M., Thiergart, L., Leech, G., Udell, D., Vazquez, J. J.,
  Mini, U., MacDiarmid, M. (2023; revised 2024). \emph{Steering Language
  Models With Activation Engineering.} arXiv:2308.10248.
\end{itemize}

\begin{center}\rule{0.5\linewidth}{0.5pt}\end{center}

\subsection{Appendix A.
Reproducibility}\label{appendix-a-reproducibility}

Frozen prompt \texttt{manager\_v2.md} (prompt-hash
\texttt{305718d8ebc3…}, \texttt{prompts/FROZEN.txt}). Preregistration
\texttt{PREREGISTRATION-v2.md} (tag \texttt{freeze-v2-2026-07-05}).
Analysis code in \texttt{final\_research/analysis/}
(\texttt{capability\_bootstrap.py}, \texttt{p2\_p3\_prereg.py},
\texttt{figures.py}, \texttt{tables.py}), all deterministic and
read-only against the decision database.

\textbf{What is released:} \texttt{data/public\_snapshot.db} --- all
21,600 decision-level outputs (stance, confidence, horizon, thesis,
risks, confession, reversal\_trigger, p\_wrong, tokens; model keys
\texttt{final-gemma4-cens}, \texttt{final-gemma4-abl-clean},
\texttt{final-qwen-cens}, \texttt{final-qwen-abl}, where \texttt{cens} =
the base/censored arm and \texttt{abl-clean} = the Gemma abliterated arm
re-collected under the base chat template, §3.2), leak-checked to
contain no upstream text; the frozen prompt; the doubt lexicon; and
every analysis script, so all tables, figures, and CIs in this paper
regenerate from public data. The dataset and analysis bundle are
permanently archived on Zenodo (DOI
\href{https://doi.org/10.5281/zenodo.21314839}{10.5281/zenodo.21314839});
code, paper, and data are also available at
\url{https://github.com/oleczek/paper-abliteration-not-a-scalpel}.
\textbf{What is not released:} the frozen
upstream bundles (analyst briefs, debate transcripts) come from a
proprietary production pipeline and are not redistributed; reproducing
the \emph{generation} step therefore requires access to that pipeline,
while reproducing every \emph{analysis} does not. Serving: BF16, vLLM,
identical flags per arm; label integrity via refusal probe; weight
hashes and config diffs recorded in \texttt{PROVENANCE.md}.

\subsection{Appendix B. Deferred / optional
analyses}\label{appendix-b-deferred--optional-analyses}

Second
abliteration author (e.g. mlabonne) for stronger method-generalization.
Mechanistic projection of the refusal direction onto a disposition axis
in activation space. A non-financial disposition battery to separate
optimism from generic positivity.

\end{document}